\documentclass[review]{elsarticle}

\usepackage{amssymb}
\usepackage{booktabs}  
\usepackage{threeparttable} 
\usepackage{multirow}
\usepackage{color}
\usepackage{hyperref}
\usepackage{amsmath}
\usepackage{graphicx}
\usepackage[table]{xcolor}
\usepackage{ulem}
\usepackage{caption}

\usepackage{stfloats}
\usepackage{makecell}
\usepackage{colortbl}
\usepackage{algorithmic}
\usepackage{algorithm}
\usepackage{pifont}
\usepackage{subfigure}
\usepackage{cleveref}
\usepackage{placeins}

\journal{Journal of Pattern Recognition}

\begin{document}

\begin{frontmatter}



\title{Time-Unified Diffusion Policy with Action Discrimination for Robotic Manipulation}

\author[1,2,3]{Ye~Niu}
\ead{niuye45678@stu.xjtu.edu.cn}
\author[1,2,3]{Sanping~Zhou\corref{mycorrespondingauthor}}
\cortext[mycorrespondingauthor]{Corresponding author: spzhou@xjtu.edu.cn}
\author[1,2,3]{Yizhe~Li}
\ead{yzl@stu.xjtu.edu.cn}
\author[4]{Ye~Deng}
\ead{dengye@swufe.edu.cn}
\author[1,2,3]{Le~Wang}
\ead{lewang@xjtu.edu.cn}

\address[1]{National Key Laboratory of Human-Machine Hybrid Augmented Intelligence}
\address[2]{National Engineering Research Center for Visual Information and Applications}
\address[3]{Institute of Artificial Intelligence and Robotics, Xi'an Jiaotong University}
\address[4]{School of Computing and Artificial Intelligence, Southwestern University of Finance and Economics}
\begin{abstract}

In many complex scenarios, robotic manipulation relies on generative models to estimate the distribution of multiple successful actions.
As the diffusion model has better training robustness than other generative models, it performs well in imitation learning through successful robot demonstrations.
However, the diffusion-based policy methods typically require significant time to iteratively denoise robot actions, which hinders real-time responses in robotic manipulation.
Moreover, existing diffusion policies model a time-varying action denoising process, whose temporal complexity increases the difficulty of model training and leads to suboptimal action accuracy.
To generate robot actions efficiently and accurately, we present the Time-Unified Diffusion Policy~(TUDP), which utilizes action recognition capabilities to build a time-unified denoising process.
On the one hand, we build a time-unified velocity field in action space with additional action discrimination information.
By unifying all timesteps of action denoising, our velocity field reduces the difficulty of policy learning and speeds up action generation.
On the other hand, we propose an action-wise training method, which introduces an action discrimination branch to supply additional action discrimination information.
Through action-wise training, the TUDP implicitly learns the ability to discern successful actions to better denoising accuracy.
Our method achieves state-of-the-art performance on RLBench with the highest success rate of 82.6\% on a multi-view setup and 83.8\% on a single-view setup.
In particular, when using fewer denoising iterations, TUDP achieves a more significant improvement in success rate.
Additionally, TUDP can produce accurate actions for a wide range of real-world tasks.
\end{abstract}




\end{frontmatter}


\section{Introduction}
\label{sec:introduction}

As an important research field in embodied intelligence, robotic manipulation has a wide range of real-world application scenarios and attracts widespread attention.
To successfully complete the robotic manipulation tasks, the next-step robot action must be predicted according to the current scene observation.
In particular, some complex tasks involve diverse successful actions, which complicates the action prediction process~\cite{jia2024towards}.
Generative models excel in this regard by effectively modeling the distribution of successful actions, leading to superior performance in various robotic manipulation tasks.
Among multiple generative models, the diffusion model has a more robust training performance in the imitation learning of robotic manipulation.
However, diffusion-based policy models take many iterations to generate the robot actions.
For example, current 3D diffusion-based policies~\cite{ke2024d,yan2024dnact} take about 100 denoising iterations, making the action denoising process consume a significant amount of time.

\begin{figure}[h]
\centering
\includegraphics[width=0.7\textwidth]{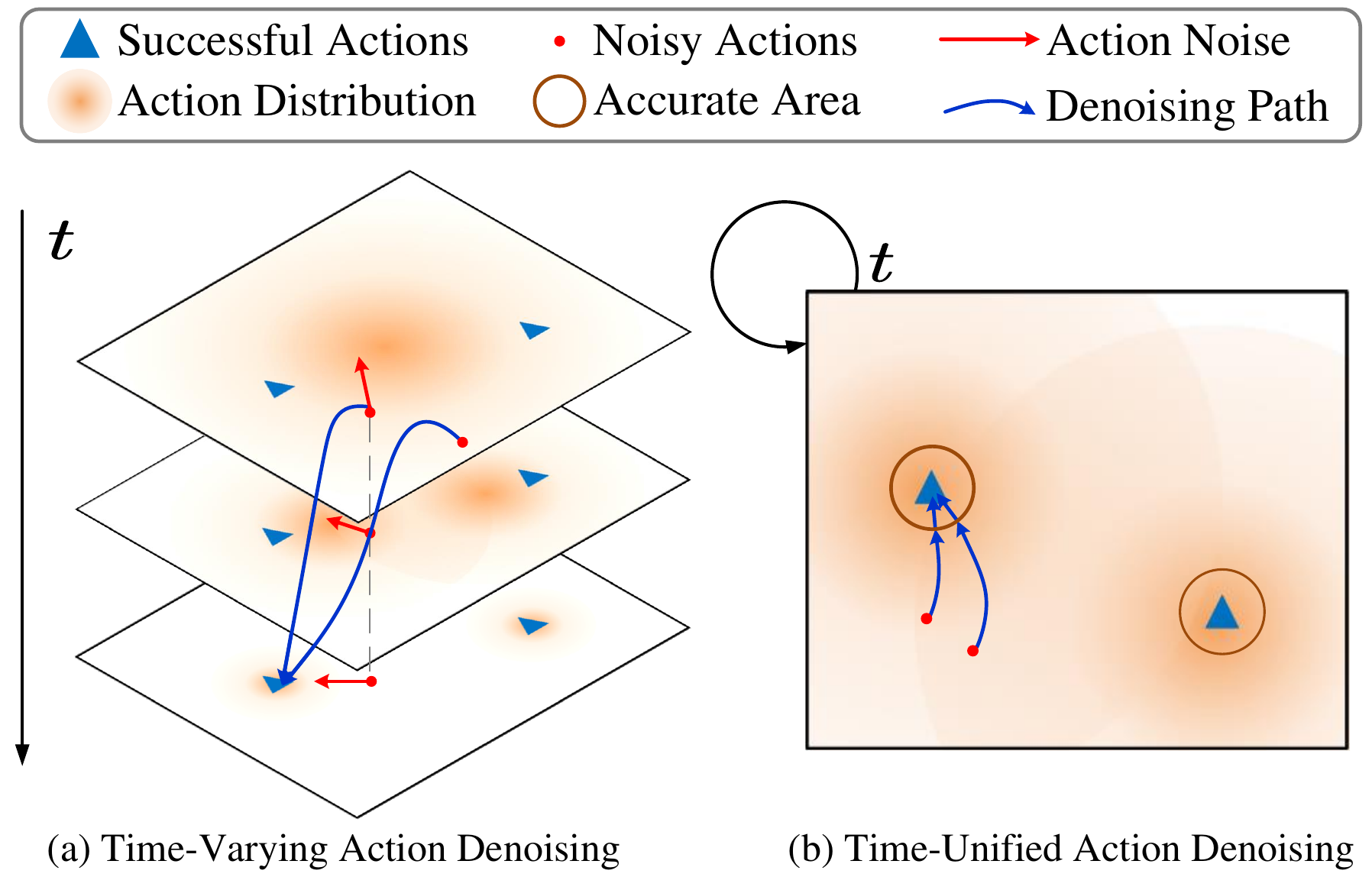} 
\caption{\textbf{Comparison of different denoising processes in action space.}
The single-step denoising directions are indicated by red arrows, and the complete denoising paths are indicated by blue arrows.
Previous diffusion policy methods adopt the time-varying velocity field, which has unclear denoising directions and high temporal complexity.
In contrast, we design the time-unified velocity field for efficient and accurate robot action generation.
}
\label{introduction}
\end{figure}

Existing diffusion policy methods have made some progress in accelerating action generation.
READ~\cite{oba2024read} designs a database of image-action pairs to retrieve better initial noise actions that are closer to successful actions, thereby reducing the number of denoising rounds.
ManiCM~\cite{lu2024manicm} obtains faster diffusion policies through consistency distillation.
FlowPolicy~\cite{zhang2025flowpolicy} uses the flow matching method to distill more efficient denoising paths.
However, the database paradigm suffers from poor task generalization, while the distillation paradigm suffers from teacher model dependence and accuracy compromise.
In order to design an action diffusion policy that balances efficiency and accuracy, we analyze two difficulties in the existing action denoising process:

1)~\textbf{Difficulty in Determining the Corresponding Successful Action.}
Since the same noisy action is related to different successful actions~\cite{ho2020denoising}, the diffusion model will be confused about which successful action to denoise.
Especially at the beginning of the iterative denoising, the noisy action approximately follows the standard Gaussian distribution, regardless of the successful actions.
As shown in Figure~\ref{introduction}(a), in the top timestep, the distributions of different successful actions after adding noise are almost identical.
Therefore, under the mutual interference of multiple successful actions, the diffusion policy cannot denoise towards a specific successful action.
As a result, the confusion between different successful actions increases time overhead and may generate inaccurate robot actions.

2) \textbf{Difficulty in Learning the Time-Varying Action Denoising.}
The diffusion policy establishes different velocity fields at different denoising timesteps, \textit{i.e.}, a time-varying denoising process.
As shown in Figure~\ref{introduction}(a), the same noisy action is mapped to different action noises at different times.
To generate accurate actions through iterative denoising, the diffusion model needs to learn the denoising ability over all denoising times.
In other words, the time-varying velocity field increases the difficulty of model training.
With limited robot demonstrations, the complexity of the time-varying velocity field indirectly reduces the accuracy of action denoising by the diffusion policy.

To address the above difficulties, we design a novel time-unified velocity field for our Time-Unified Diffusion Policy~(TUDP) to develop the denoising process.
On the one hand, TUDP builds a time-unified velocity field with low timing complexity for the network to fit.
Furthermore, our velocity field emphasizes the neighborhood of successful actions, within which straight paths are modeled to guarantee denoising accuracy, as shown in Figure~\ref{introduction}(b).
As a result, TUDP could denoise the action space to successful actions with fewer denoising iterations.
On the other hand, TUDP proposes an action-wise training method for TUDP with two training steps.
Firstly, we introduce and train an action discrimination network to discriminate successful actions.
Secondly, we design a novel action-weighted loss function, optimizing the unified diffusion network in conjunction with action discrimination information.

We evaluate our method on RLBench simulation tasks, which achieves state-of-the-art performance with the highest average success rate.
The real-robot experiments also shows the effectiveness of our method.
In addition, we verify the justification of the components of our method through ablation experiments.

In summary, our main contributions are as follows:
\begin{itemize}

\item We design a novel TUDP for efficient and accurate action denoising.
Our action policy gains action discrimination capabilities to clear the denoising directions.

\item We design the time-unified velocity field with additional action discrimination information to efficiently converge the action space to successful actions, which avoids conflicts at different timesteps.

\item We propose the action-wise training method to incorporate motion discrimination into denoising capabilities, through action discrimination training and the action-weighted loss function jointly.

\end{itemize}
The following paper is organized as follows: Section~\ref{Related Work} briefly reviews some related works. In Section~\ref{Method}, we introduce the network architecture and training method of our TUDP in detail. Section~\ref{Experiments} presents performance comparisons and ablation experiments in both simulation and real-world environments, and the conclusion is summarized in Section~\ref{Conclusion}.

\section{Related Work}
\label{Related Work}
\textbf{Diffusion Model.}
Through iterative denoising processes, early diffusion models~\cite{ho2020denoising,song2021denoising} enable diverse and high-quality visual generation.
Since the diffusion process could be modeled as a stochastic differential equation~\cite{song2021scorebased}, the continuous diffusion models~\cite{dockhorn2022scorebased,jolicoeurmartineau2021gotta} achieve more efficient generation with fewer steps.
To achieve higher computational efficiency, the widely-regarded stable diffusion adopts a latent space~\cite{rombach2022high,vahdat2021score,davtyan2023efficient}, while the lower bound on the dimension of the latent space is still limited by image decoding.
In the last few years, recent works have provided a deeper analysis of diffusion models to improve the generation results.
Cold Diffusion~\cite{bansal2024cold} designs a more robust iteration to revert arbitrary degradation.
Flow Matching~\cite{lipman2022flow} build simple denoising paths, reducing reliance on large amounts of time consumption.
Some recent works~\cite {lin2024common,zhang2024tackling} have noticed and attempted to address the subtle differences in distributions between training and inference.
Research on several samples~\cite{wu2024cgi} utilizes sufficient fine-tuning to facilitate model training.

\textbf{Imitation Learning from Robot Demonstrations.}
Imitation learning has gained significant traction as a powerful framework for robotic manipulation tasks, allowing robots to navigate the intricacies of manipulation tasks without explicitly programmed rules or rewards~\cite{zeng2021transporter,zare2024survey,lu2024manigaussian}.
Recently, various methods have been developed for learning manipulation policies with different task constraints and control modalities. 
Action chunking transformer~\cite{zhao2023learning} demonstrated the effectiveness of imitation learning in bi-manual manipulation tasks. 
RVT~\cite{goyal2023rvt} performs well in language-conditioned tasks based on imitation learning, both in the RLBench environment and in real-world scenarios. 
Some research~\cite{ho2016generative,tsurumine2022goal} applies generative models to imitation learning, such as energy-based models~\cite{florence2022implicit,ta2022conditional}.
As diffusion models progressively convert a basic prior distribution into a target distribution, existing works~\cite{ryu2024diffusion,urain2023se,pearce2023imitating} model state-conditioned action distributions in imitation learning.
Particularly, the diffusion policy~\cite{chi2023diffusion} generates actions by imitating demonstrations. 

\textbf{Diffusion Policy in Robotic Manipulation.} 
In recent years, a series of works have verified the potential of diffusion models in robotic manipulation.
Different from using diffusion models to generate more visual scene information~\cite{wu2024unleashing}, the potential of diffusion models to predict actions has also been explored.
Diffusion Policies~\cite{DBLP:journals/corr/abs-2303-04137,ze20243d} successfully models the probability of trajectory sequences in different tasks.
3D Diffusion Policy~\cite{ze2024d} incorporates the power of 3D visual representations into conditional diffusion models.
DNActor~\cite{yan2024dnact} distills 2D semantic features from foundation models, such as Stable Diffusion~\cite{rombach2022high} and NeRF~\cite{driess2022reinforcement}, to a 3D space in its pretrain phase.
Flow-based diffusion policies~\cite{chisari2024learning,DBLP:journals/corr/abs-2409-04576} generate robot actions via more direct denoising paths.
With the continuous improvement of diffusion models in the field of visual generation, works develop the diffusion paradigms in robotic manipulation.
Hierarchical diffusion policy \cite{ma2024hierarchical} adds robot kinematic constraint on the diffusion models.
READ~\cite{oba2024read} preserves the kinematic feasibility of the generated action via forward diffusion in a low-dimensional latent space, while using cold diffusion to achieve high-resolution action via back diffusion in the original task space.
These methods inspire us to enhance the diffusion policy for action denoising.

\begin{figure*}[t]   
	\centering
	\includegraphics[width=\linewidth]{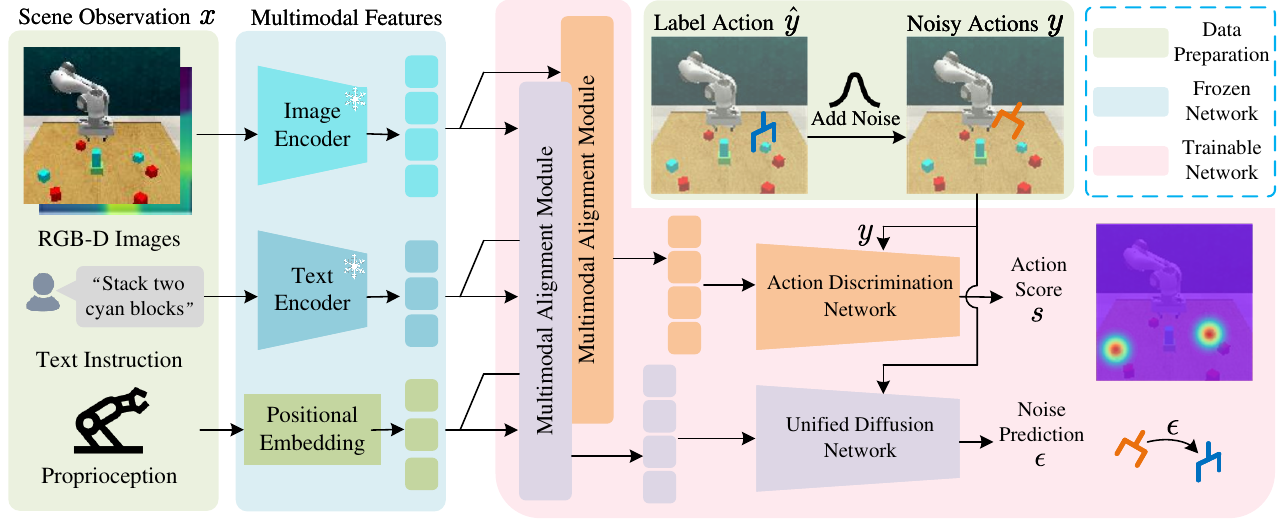}
	\caption{\textbf{The architecture of TUDP.}
    The action score $s$ is used to identify successful actions.
    And the noise prediction $\epsilon$ is used to correct noisy action $y$. }
	\label{overview}
\end{figure*}

\section{Method}
\label{Method}
In this section, we propose the Time-Unified Diffusion Policy~(TUDP) for efficient and accurate robotic manipulation.
The overall framework of TUDP is illustrated in Figure~\ref{overview}.
We first describe the keyframe-based robotic manipulation and the data flow process in our network.
We then delve into the main development of our approach, including the time-unified velocity field and action-wise training method.

\subsection{Overview}

Our TUDP adopts a keyframe-based manipulation framework, which selects discrete keyframes to compose the robot demonstrations~\cite{james2022coarse}.
Essentially, each keyframe records the label action $\hat{y}$ and scene observation $x$, which contains multi-view RGB-D images, the text instruction, and historical robot actions named proprioception. 
Since the robot interaction depends directly on the end effector rather than joint angles, we use end-effort posture $y$ to guide the robot actions, which includes the translation $y_{pos}$, the rotation $y_{rot}$, and the binary opening state $y_{open}$ of the gripper, as shown below:
\begin{equation}
    \label{action parts}
    y = \{y_{pos}\! \in \! \mathbb{R}^3,~y_{rot}\! \in \! SO(3),~y_{open}\! \in \! \{0,1\}\}.
\end{equation}
Since the opening states contain little spatial location information, our network discards the opening states of input noisy actions.
In successful robot demonstrations during training, each keyframe scene $x$ corresponds to a label action $\hat{y}$, which is one of the successful actions $\{\hat{y}^i\}_{i=1}^k$.

\newcommand{\myComment}[1]{\textcolor{teal}{~~\% #1}}

\begin{algorithm}[t]
    \caption{Time-Unified Iterative Denoising}
    \label{alg:inference}
    \begin{algorithmic}[1]
        \STATE \textbf{Input:} Scene observation $x$, noisy action $y_0$, number of iterations $N$
        \STATE \textbf{Output:} Denoised action $y_N$
        \STATE \textbf{Initialize:} Network $\epsilon_\vartheta$(x,y) for noise prediction
        
        \FOR{$t = 0,1,...,N$} 
            \STATE Compute the predicted noise: $\epsilon = \epsilon_\vartheta(x, y_t)$
            \STATE Update the noisy action:
            \(
            y_{t+1} =  y_t - \epsilon\)
        \IF{$\|y_{t+1}-y_t\|<\delta$}
                \STATE $y_N = y_{t+1}$
                \STATE Break 
                \myComment{Early termination}
            \ENDIF
        \ENDFOR
        \STATE \textbf{Return} $y_N$ 
    \end{algorithmic}
\end{algorithm}

The network architecture of TUDP is shown in Figure~\ref{overview}.
We acquire multimodal features in a training-free manner using pretrained models.
In particular, we use a CLIP image encoder for multi-view images and a CLIP text encoder for text instructions.
Besides, the proprioception including historical robot actions is processed by rotary position embedding.
Then the multimodal features are fed into two parallel multimodal alignment modules to align and fuse multimodal features.
The two branches are connected to the action discrimination network and the unified diffusion network respectively, whose functions are analyzed below.

\textbf{Action Discrimination Network.}
In order to construct the time-unified velocity field during training, we propose the action discrimination network.
The action discrimination network predicts the action score $s$ of the noisy action $y$, which only functions during the training phase.
In fact, the action discrimination network determines whether the noisy action $y$ is in the neighborhood $U(\hat{y}^i,l)$ of a successful action $\hat{y}^i$ with radius $l$.
Specifically, we hope the label $\hat{s}$ of action score to be 1 where $\|y-\hat{y}\|<l$ and 0 where $\|y-\hat{y}\|>l$.
Assuming that successful actions are sparse in common scenarios, we could choose a tiny neighborhood radius $l$ to avoid the overlap of different neighborhoods.
Finally, the heatmap of the label $\hat{s}$ in the action space is shown on the right side of Figure~\ref{overview}.

\textbf{Unified Diffusion Network.}
The time-unified diffusion network takes scene observations and noisy actions to predict action noises, unlike other diffusion policy networks~\cite{ke2024d, oba2024read} acquiring denoising time $t$ as input.
As noisy actions in the action space are sampled with probability $p(y|\hat{y})$, the unified diffusion network is expected to denoise arbitrary noisy actions after sufficient training.
In other words, the unified diffusion network learns a time-unified velocity field in the action space according to the scene observation.
Compared with the time-varying velocity field $\epsilon(y,t)$ modeled by the diffusion model, the time-unified velocity field $\epsilon(y)$ is easier to fit by the network.
As shown in Algorithm~\ref{alg:inference}, we propose time-unified iterative denoising based on the unified diffusion network.
Furthermore, we design the early termination to flexibly shorten iterations, where $\delta=0.01$ is the condition for early termination.


\begin{figure}[!t]\centering
	\includegraphics[width=8cm]{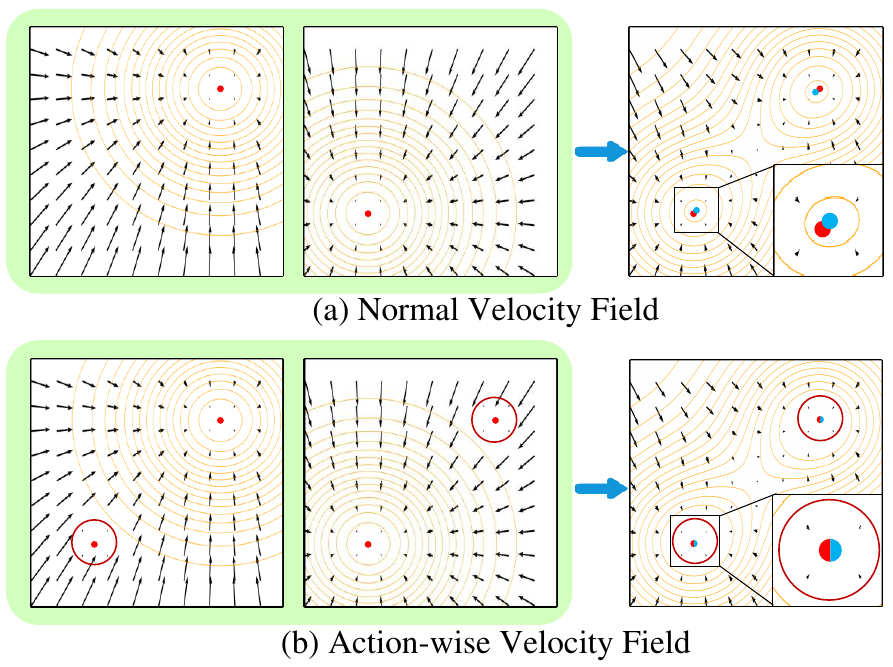}
	\caption{\textbf{Comparison of two time-unified velocity fields.} 
    Red dots denote successful actions, blue dots denote denoised actions, half-red and half-blue dots indicate that the denoised actions equal the successful actions, and the orange line is the probability contour of the noisy action.}
    \label{expect field}
\end{figure}

\subsection{Time-unified Velocity Field}

We design a time-unified velocity field to denoise the action space towards successful actions, which is constant during the iterative denoising process.
Since only one of the successful actions is labeled in the scene during training, without additional information, we can just obtain the conditional velocity field $\varepsilon(y|\hat{y})$. 
Considering scenarios with a single successful action, the conditional velocity field could point to the successful action with a simple implementation as $\varepsilon(y|\hat{y})=y-\hat{y}$.
To constrain the range of action noise, we add a limitation $v$ on the conditional velocity field, as shown below:
\begin{equation}
    \label{conditional field}
    \varepsilon(y|\hat{y}) = v\cdot \frac{y-\hat{y}}{\max\{v,\|y-\hat{y}\|\}},
\end{equation}
Then, as shown in Figure~\ref{expect field}(a), the conditional velocity fields are merged into the overall velocity field $\varepsilon(y)$ as follows:
\begin{equation}
    \label{global velocity field}
    \varepsilon(y) = \mathbb{E}_{p(\hat{y}^i)p(y|\hat{y}^i)} \varepsilon(y|\hat{y}^i),
\end{equation}
in which the action sampling distribution $p(y|\hat{y}^i)$ is greater than $0$ everywhere in action space to cope with random initialized noisy action, and the successful action distribution $p(\hat{y}^i)$ equals $\frac{1}{k}$ as common.
However, the constant velocity field $\varepsilon(y)$ mostly corresponds to non-zero noises at successful actions as follows:
\begin{equation}
    \label{non-zero bias}
    \varepsilon(\hat{y}^i) = \sum_{j\neq i} \frac{1}{k}\cdot p(\hat{y}^i|\hat{y}^j)\cdot \varepsilon(\hat{y}^i|\hat{y}^j) \neq 0,
\end{equation}
which means that the predicted actions from iterative denoising deviate from the successful actions.

Since the deviation is essentially caused by the interference of other successful actions $\{\hat{y}^j\}_{j\neq i}$ on the desired successful action $\hat{y}^i$, 
we introduce additional action correlation information. 
Specifically, the correlation weight $\lambda(y,\hat{y}^i)$ depends on the minimum distance between the noisy action and other successful actions as follows:
\begin{equation}
    \label{lambda}
    \lambda(y,\hat{y}^i) = \left\{\begin{array}{rcl} 0 & \mbox{for} &\min_{j\neq i} \|y-\hat{y}^j\|\leq l,\\
    1 & \mbox{for} &\min_{j\neq i} \|y-\hat{y}^j\|>l,\end{array}\right.
\end{equation}
in which hyperparameter $l$ defines the neighborhood radius of successful actions.
By utilizing correlation weight $\lambda$, we design the conditional time-unified velocity field based on label action $\hat{y}$ as follows:
\begin{equation}
    \label{new conditional}
    \epsilon(y|\hat{y})=\lambda(y,\hat{y})\varepsilon(y|\hat{y}).
\end{equation}
Afterward, since $k$ successful actions have the same probability of becoming the label action, conditional velocity fields $\epsilon(y|\hat{y}^i)$ compose the time-unified velocity field $\epsilon(y)$ in Figure~\ref{expect field}(b) expressed as follows:
\begin{equation}
    \label{new merge of field}
    \epsilon(y) = \sum_{i=0}^k \frac{1}{k}\cdot p(y|\hat{y}^i)\lambda(y,\hat{y}^i) \varepsilon(y|\hat{y}^i).
\end{equation}

\begin{algorithm}[t]
    \caption{Action-wise Training Process}
    \begin{algorithmic}[1]
        \STATE \textbf{Dateset:} Scene observations \{$x$\}, label actions $\{\hat{y}\}$
        \STATE \textbf{Model:} Action discrimination network $s_\theta$, unified diffusion network $\epsilon_\vartheta$
        \STATE \textbf{Initial:} Distribution of noisy actions $p(y|\hat{y})$, maximum training round $D$
        \STATE \textbf{Phase 1: Action discrimination training}
        \FOR{rounds from 1 to $D$}
                \STATE Sample $(x,\hat{y})$ and $y\in p(y|\hat{t})$
                \STATE Calculate the action score label $\hat{s}$
                \STATE Optimize $\theta$ with loss $\mathcal{L}_{score}(s_\theta(x,y),\hat{s})$ 
            \ENDFOR
        \STATE \textbf{Phase 2: time-unified diffusion training}
        \FOR{rounds from 1 to $D$}
                \STATE Sample $(x,\hat{y})$ and $y\in p(y|\hat{y})$
                \STATE Calculate action weights~$\lambda(s,y,\hat{y})$ with $s_\theta(x,y)$
                \STATE Calculate conditional velocity field $\varepsilon(y|\hat{y})$
                \STATE Optimize $\vartheta$ with loss $\mathcal{L}_{noise}(\epsilon_\vartheta(x,y),\varepsilon,\lambda)$ 
            \ENDFOR
        
        \STATE \textbf{Return} model $s_\theta$, $\epsilon_\vartheta$
    \end{algorithmic}
    \label{alg:train}
\end{algorithm}

Consequently, the time-unified velocity field $\epsilon(y)$ makes the action space converge to successful actions. 
When noisy actions are samples outside of the neighborhoods of successful actions, appropriate conditional distribution $p(y|\hat{y})$ can construct the time-unified velocity field to denoise noisy actions into the neighborhoods.
In the neighborhood of a random successful action $\hat{y}^i$, the time-unified velocity field $\epsilon(y)$ is ideally equal to $\epsilon(y|\hat{y}^)$, which denoise $y\in U(\hat{y}^i,l)$ to successful action $\hat{y}^i$.
In short, the neighborhood of successful actions can converge to the corresponding successful action in one step under the guidance of the time-unified velocity field.

\begin{figure}[!t]\centering
	\includegraphics[width=8.5cm]{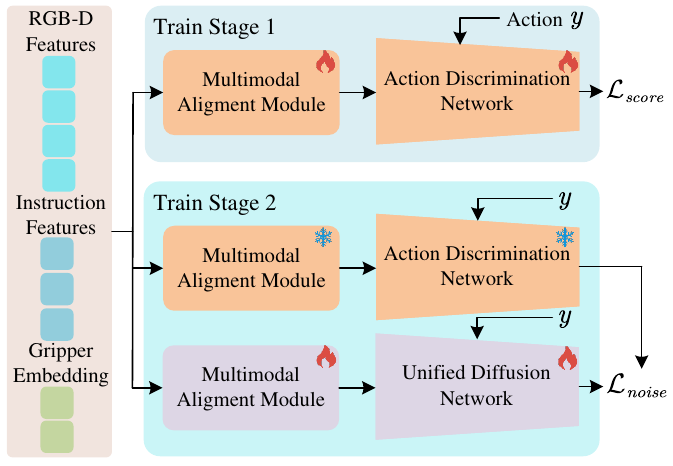}
	\caption{\textbf{Action-wise training method with two stages.} 
    We omit the multimodal encoders in our method that are training-free.}
    \label{fig:train}
\end{figure}

\subsection{Action-wise Training Method}
\label{section:train}

As shown in Figure~\ref{fig:train}, we propose the action-wise training method to optimize the action discrimination network and the unified diffusion network separately.
Since robot demonstrations for training include scene observations and label actions in pairs, the noisy actions are sampled from the conditional distribution as follows:
\begin{equation}
    \label{conditional p}
    p(y|\hat{y}) = \frac{1}{\sqrt{2\pi \sigma^2}} e^{-\frac{(y - \hat{y})^2}{2\sigma^2}},
\end{equation}
in which $\sigma$ is a hyperparameter to balance the global action space and the local successful action.
The two-stage training process of the action-wise training method is shown in Algorithm~\ref{alg:train}.

In the first stage, we train the action discrimination network, which distinguishes between the successful action neighborhoods and the outside regions of them.
In order to avoid learning errors caused by the sensitivity near the boundary $\|y-\hat{y}\|=l$, we adopt the following continuous negative exponential function:
\begin{equation}
    \label{s}
    \hat{s}(y,\hat{y}) = e^{m\cdot \text{ReLU}(\|y-\hat{y}\|-l)},
\end{equation}
in which $m$ is set to be less than $\text{-}10$. 
Then the loss function of the action score prediction $s$ is as follows:
\begin{equation}
    \label{loss of action network}
    \mathcal{L}_{score} = \mathbb{E}_{p(x,\hat{y})p(y|\hat{y})} \|s_\theta(x,y)-\hat{s}(y,\hat{y})\|.
\end{equation}

In the second stage, we train the unified diffusion network $\epsilon_\vartheta(x,y)$ to learn the time-unified velocity field $\epsilon_x(y)$ based on scene $x$.
Based on action score $s_\theta(x,y)$ from the action discrimination network, we use the following formula to approximate the correlation weight $\lambda$:
\begin{equation}
    \lambda(y,\hat{y}) = 1-s_\theta(x,y) \cdot \text{sgn}(\|\hat{y}-y\|-l).
\end{equation}
Upon the training demonstrations with label actions, the action loss is formulated as follows:
\begin{align}
    \label{noise loss}
    \mathcal{L}_{action} &=\mathbb{E}_{p(x,y)}\|\epsilon_\vartheta(x,y)-\epsilon_x(y)\| \\
    &= \mathbb{E}_{p(x,\hat{y})p(y|\hat{y})} \|\epsilon_\vartheta(x,y)-\lambda(y,\hat{y})\varepsilon(y|\hat{y})\|
\end{align}
Finally, we show the overall loss $\mathcal{L}_{noise}$ of the second training stage below:
\begin{equation}
    \label{overall loss}
    \mathcal{L}_{noise} = \mathcal{L}_{action} + w_{open}\mathcal{L}_{open}
\end{equation}
where the opening weight $w_{open}=0.4$ and the gripper opening loss $\mathcal{L}_{open}$ is as follows:
\begin{equation}
    \label{open loss}
    \small
    \mathcal{L}_{open} = y_{\text{open}}\log(\hat{y}_{\text{open}})-(1-y_{\text{open}})\log(1-\hat{y}_{\text{open}}).
\end{equation}

\begin{figure*}[t]
\centering
\includegraphics[width=1\textwidth]{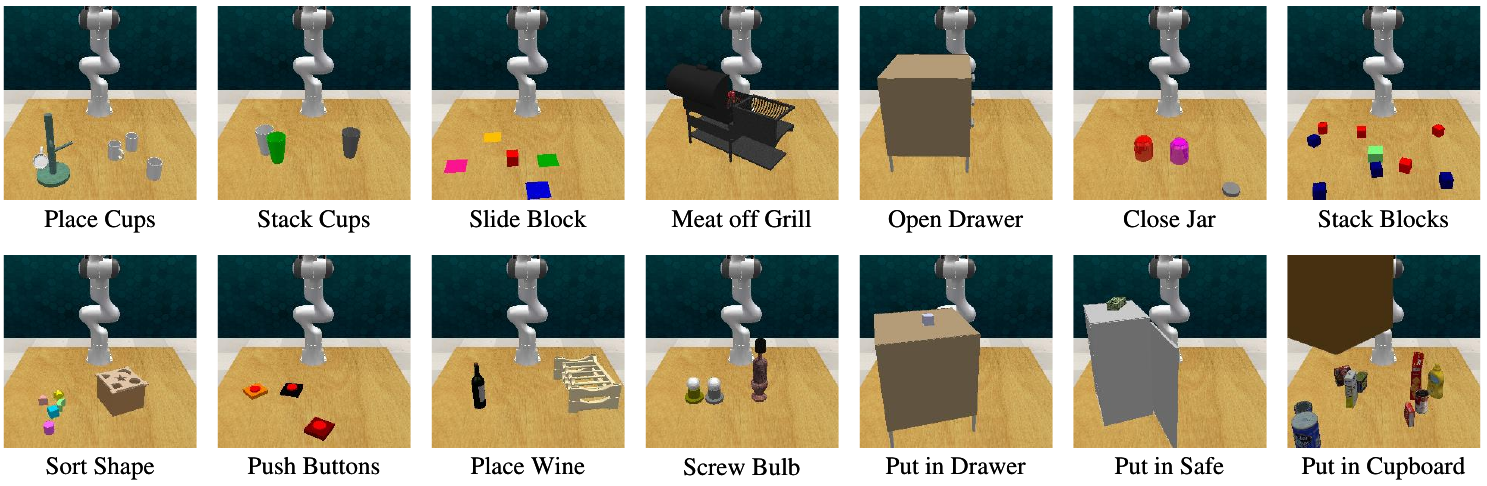} 
\caption{\textbf{Overview of 14 tasks.} We show the image observations in these RLBench tasks.}
\label{demonstrations}
\end{figure*}

\section{Experiments}
\label{Experiments}
\renewcommand{\arraystretch}{0.7} 
\begin{table*}[t]
	\centering  
    \setlength{\tabcolsep}{3.pt}
	\fontsize{6.5}{9}\selectfont  
		\begin{tabular}{c|ccccccc}
		\toprule
			Models&\makecell{Avg.\\[-7pt] Success.}&\multicolumn{1}{c|}{\makecell{Avg.\\[-7pt] Rank.}}&\makecell{Place\\[-7pt] Cups}&\makecell{Stack\\[-7pt] Cups}&\makecell{Sort\\[-7pt] Shape}&\makecell{Push\\[-7pt] Buttons}&\makecell{Stack\\[-7pt] Blocks}\cr 
			\hline
            PolarNet&46.4& \multicolumn{1}{c|}{6.3}&0.0&8.0&12.0&96.0&4.0\cr
			PerAct&49.4&\multicolumn{1}{c|}{6.2}&$2.4^{\pm 3.2}$&$2.4^{\pm 2.2}$&$16.8^{\pm 4.7}$&$92.8^{\pm 3.0}$&$26.4^{\pm 3.2}$\cr
            HiveFormer&47.3&\multicolumn{1}{c|}{6.7}&0.0&0.0&8.0&84.0&8.0\cr
            Act3D&63.2&\multicolumn{1}{c|}{4.8}&$3.2^{\pm 3.0}$&$9.6^{\pm 6.0}$&$29.6^{\pm 3.2}$&$93.6^{\pm 2.0}$&$4.0^{\pm 3.6}$\cr
            RVT&62.9&\multicolumn{1}{c|}{4.6}&$4.0^{\pm 2.5}$&$26.4^{\pm 8.2}$&$36.0^{\pm 2.5}$&$\textbf{100}^{\pm 0.0}$&$28.8^{\pm 3.9}$\cr
            RVT2&81.4&\multicolumn{1}{c|}{2.4}&$\textbf{38.0}^{\pm 4.5}$&$\textbf{69.0}^{\pm 5.9}$&$35.0^{\pm 7.1}$&$\textbf{100.0}^{\pm 0.0}$&$\textbf{80.0}^{\pm 2.8}$\cr
            3D Diffuser Actor&81.3&\multicolumn{1}{c|}{2.4}&$24.0^{\pm 7.6}$&$47.2^{\pm 8.5}$&$\underline{44.0}^{\pm 4.4}$&$98.4^{\pm 2.0}$&$68.3^{\pm 3.3}$\cr
            \rowcolor{gray!20}
            \textbf{TUDP~(Ours)}&\textbf{82.6}&\multicolumn{1}{c|}{\textbf{1.9}}&$\underline{32.0}^{\pm 5.2}$&$\underline{53.0}^{\pm 3.3}$&$\textbf{48.0}^{\pm 3.8}$&$98.0^{\pm 2.0}$&$\underline{69.0}^{\pm 3.8}$\cr
            \hline
            \hline
            Models&\makecell{Open\\[-7pt] Drawer}&\makecell{Close\\[-7pt] Jar}&\makecell{Place\\[-7pt] Wine}&\makecell{Screw\\[-7pt] Bulb}&\makecell{Put in\\[-7pt] Drawer}&\makecell{Put in\\[-7pt] Safe}&\makecell{Drag\\[-7pt] Stick}\cr
            \hline
            PolarNet&84.0&36.0&40.0&44.0&32.0&84.0&92.0\cr
            PerAct&$88^{\pm 5.7}$&$55.2^{\pm 4.7}$&$44.8^{\pm 7.8}$&$17.6^{\pm 2.0}$&$51.2^{\pm 4.7}$&$84^{\pm 3.6}$&$89.6^{\pm 4.1}$\cr
            HiveFormer&52.0&52.0&80.0&8.0&68.0&76.0&76.0\cr
            Act3D&$78.4^{\pm 11.2}$&$96.8^{\pm 3.0}$&$59.2^{\pm 9.3}$&$32.8^{\pm 6.9}$&$91.2^{\pm 6.9}$&$95.2^{\pm 4.0}$&$80.8^{\pm 6.4}$\cr
            RVT&$71.2^{\pm 6.9}$&$52.0^{\pm 2.5}$&$91.0^{\pm 5.2}$&$48.0^{\pm 5.7}$&$88.0^{\pm 5.7}$&$91.2^{\pm 3.0}$&$99.2^{\pm 1.4}$\cr
            RVT2&$74.0^{\pm 11.8}$&$\textbf{100.0}^{\pm 0.0}$&$\underline{95.0}^{\pm 3.3}$&$\textbf{88.0}^{\pm 4.9}$&$\textbf{96.0}^{\pm 0.0}$&$96.0^{\pm 2.8}$&$99.0^{\pm 1.7}$\cr
            3D Diffuser Actor&$\underline{89.6}^{\pm 4.1}$&$96.0^{\pm 2.5}$&$93.6^{\pm 4.8}$&$\underline{82.4}^{\pm 2.0}$&$\textbf{96.0}^{\pm 3.6}$&$\underline{97.6}^{\pm 2.0}$&$\textbf{100}^{\pm 0.0}$\cr
            \rowcolor{gray!20}
            \textbf{TUDP~(Ours)}&$\textbf{93.0}^{\pm 2.0}$&$\textbf{100.0}^{\pm 0.0}$&$\textbf{96.0}^{\pm 3.2}$&$80.0^{\pm 3.2}$&$\textbf{96.0}^{\pm 2.3}$&$\textbf{100.0}^{\pm 0.0}$&$\textbf{100.0}^{\pm 0.0}$\cr
            \hline
            \hline
        \end{tabular} 
        \begin{tabular}{c|ccccccc}
            Models&\makecell{Put in\\[-7pt] Cupboard}&\makecell{Slide\\[-7pt] Block}&\makecell{Meat off\\[-7pt] Grill}&\makecell{Insert\\[-7pt] Peg}&\makecell{Sweep to\\[-7pt] Dustpan}&\makecell{Turen\\[-7pt] Tap}\cr
            \hline
            PolarNet&12.0&56.0&100.0&4.0&52.0&80.0\cr
            PerAct&$28^{\pm 4.4}$&$74^{\pm 13.0}$&$70.4^{\pm 2.0}$&$5.6^{\pm 4.1}$&$52^{\pm 0.0}$&$88^{\pm 4.4}$\cr
            HiveFormer&68.0&64.0&\textbf{100}&0.0&28.0&80.0\cr
            Act3D&$67.2^{\pm 3.0}$&$96.0^{\pm 2.5}$&$95.2^{\pm 1.6}$&$24.0^{\pm 8.4}$&$86.4^{\pm 6.5}$&$94.4^{\pm 2.0}$\cr
            RVT&$49.6^{\pm 3.2}$&$81.6^{\pm 5.4}$&$88^{\pm 2.5}$&$11.2^{\pm 3.0}$&$72.0^{\pm 0.0}$&$93.6^{\pm 4.1}$\cr
            RVT2&$66.0^{\pm 4.5}$&$92.0^{\pm 2.8}$&$99.0^{\pm 1.7}$&$40.0^{\pm 0.0}$&$\textbf{100.0}^{\pm 0.0}$&$\underline{99.0}^{\pm 1.7}$\cr
            3D Diffuser Actor&$\textbf{85.6}^{\pm 4.1}$&$\underline{97.6}^{\pm 3.2}$&$96.8^{\pm 1.6}$&$\textbf{65.6}^{\pm 4.1}$&$84^{\pm 4.4}$&$\textbf{99.2}^{\pm 1.6}$\cr
            \rowcolor{gray!20}
            \textbf{TUDP(Ours)}&$\underline{76.0}^{\pm 2.3}$&$\textbf{100.0}^{\pm 0.0}$&$98.00^{\pm 2.0}$&$\underline{54.0}^{\pm 5.2}$&$\underline{97.0}^{\pm 2.0}$&$97.0^{\pm 2.0}$\cr
            \bottomrule
		\end{tabular} 
	\caption{\textbf{Evaluation on RLBench with multiple camera views}. Our approach achieved the highest average task success rate. Black bold fonts indicate the best performance and underlines indicate suboptimal performance for each task.}
    \label{tab:performance_comparison} 
\end{table*}

\subsection{Experimental Setup}


\textbf{Dataset and Simulation.} 
We train and evaluate our TUDP on a multi-task manipulation benchmark developed in RLBench~\cite{james2020rlbench} with a 7-DOF Franka Panda Arm. 
We use 18 distinct tasks, each comprising 150 demonstrations, with 14 tasks illustrated in Figure~\ref{demonstrations}.
Each task includes various text instructions, ranging from 2 to 60 variations. 
These variations often include different requirements for locations and colors.
The demonstrations are collected in a simulation environment built by CoppeliaSim~\cite{rohmer2013v}.
The $256\times256$ RGB-D images in demonstrations are captured by four noiseless cameras positioned at the front, left shoulder, right shoulder, and wrist of the robot.
In the simulation environment, the robot arm has an accurate proprioception of its own state.

\textbf{Training and Evaluation Details.}
Our TUDP is trained on 4 NVIDIA 3090Ti 10GB GPUs for $\sim$40K steps with a cosine learning rate decay schedule. 
We adopt a batch size of 32 and initialize the learning rate to $10^{\text{-}4}$.
Among the 150 pre-generated demonstrations, 100 are used for training, 25 for validation, and 25 for testing. 
To be consistent with other methods, we tested each task four times in the multi-view setting to calculate the average success rates and their Gaussian variance.
We compared the performance of our method with existing methods in both simulation experiments and real machine experiments.

\textbf{Baselines.}
On the RLBench benchmark, we compare TUDP with the previous state-of-the-art methods, which achieve excellent performance.
The following work has improved the accuracy of robotic manipulation strategies: PolarNet~\cite{chen23polarnet}, HiveFormer~\cite{guhur2023instruction}, PerAct~\cite{shridhar2023perceiver}, Act3D~\cite{gervet2023act3d}, RVT~\cite{goyal2023rvt}.
Additionally, RVT2~\cite{DBLP:journals/corr/abs-2406-08545} using action-value graphs and 3D Diffuser Actor~\cite{ke2024d} using the diffusion model make progress in modeling multiple successful actions.

\begin{table}
    \centering
    \setlength{\tabcolsep}{2pt}
    \fontsize{8.5}{12}\selectfont  
    \begin{tabular}{c|cccccc}
    \toprule
    Models& \makecell{Avg.\\[-5pt] Success} & \multicolumn{1}{c|}{\makecell{Avg.\\[-5pt] Rank.}} & \makecell{close\\[-5pt] jar}& \makecell{open\\[-5pt] drawer}& \makecell{sweep to\\[-5pt] dustpan}& \makecell{meat off\\[-5pt] grill}\cr
    \hline
    GNFactor&31.7&\multicolumn{1}{c|}{4.0}&25.3&76.0&28.0&57.3\cr
    Act3D&65.3&\multicolumn{1}{c|}{2.8}&52.0&84.0&80&66.7\cr
    3D Diffuser Actor&78.4&\multicolumn{1}{c|}{1.8}&\textbf{82.7}&\textbf{89.3}&94.7&88.0\cr
    \rowcolor{gray!20}
    \textbf{TUDP~(Ours)}&\textbf{83.8}&\multicolumn{1}{c|}{1.4}&78.0&88.0&\textbf{98.0}&\textbf{92.0}\cr
    \hline
    \hline
    Model& \makecell{turn\\[-5pt] tap}& \makecell{slide\\[-5pt] block}& \makecell{put in\\[-5pt] drawer}&\makecell{drag\\[-5pt] stick}& \makecell{push\\[-5pt] buttons}& \makecell{stack\\[-5pt] blocks}\cr
    \hline
    GNFactor&50.7&20.0&0.0&37.3&18.7&4.0\cr
    Act3D&64.0&\textbf{100.0}&54.7&86.7&64.0&0.0\cr
    3D Diffuser Actor&80.0&92.0&77.3&98.7&69.3&12.0\cr
    \rowcolor{gray!20}
   \textbf{TUDP~(Ours)}&\textbf{85.0}&90.0&\textbf{91.0}&\textbf{100.0}&\textbf{96.0}&\textbf{21.0}\cr
    \bottomrule
    \end{tabular}
    \caption{\textbf{Evaluation on RLBench with single camera view.} We report average success rates on 10 RLBench tasks with only the \textit{front} camera view. }
    \label{Evaluation on single view camera}
\end{table}

\subsection{Simulation Results}

 \begin{figure}[t]
    \centering  
    \includegraphics[width=7cm]{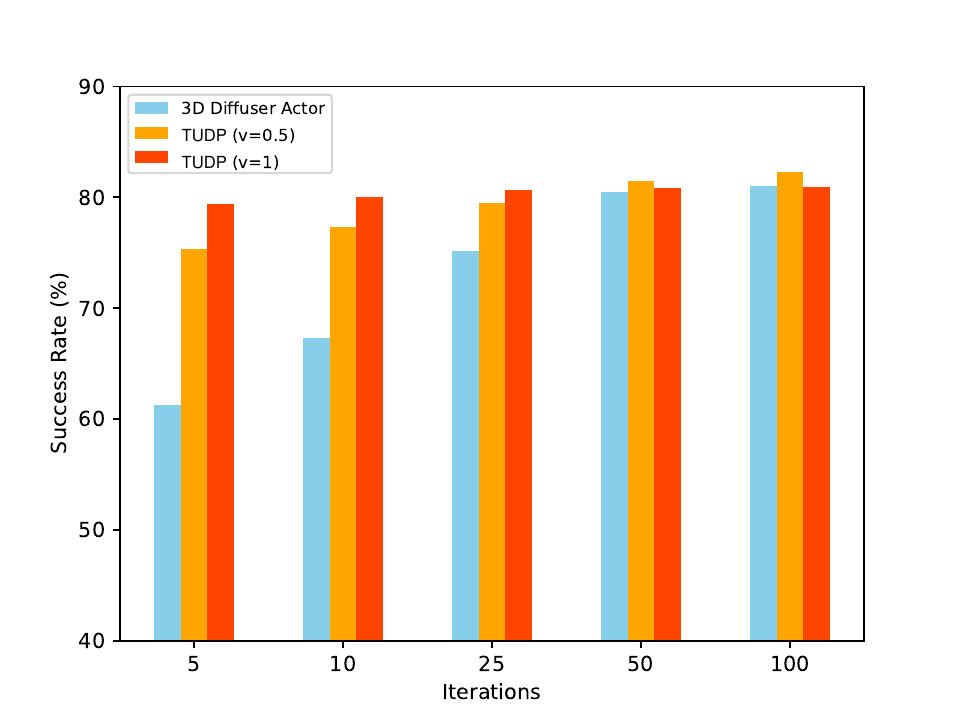}  
    \caption{\textbf{Comparison of average success rates at different iterations.}
    When using fewer denoising iterations, our TUDP achieves a significantly higher success rate than the 3D Diffuser Actor.}
    \label{comparison of efficiency and accuracy}
 \end{figure}

\textbf{Success Rate in a Multi-view Setting.}
Following the setting of PerAct~\cite{shridhar2023perceiver}, we conduct experiments on 18 tasks with 4 camera views.
For each task, we conducted four repeated experiments to calculate the average task success rate and standard deviation.
These 18 tasks cover a wide range of robot action types and task requirements, and can comprehensively reflect the performance of robot action policies.
As shown in Table~\ref{tab:performance_comparison}, we calculated the average success rate and average performance ranking on 18 tasks.
Our TUDP achieves the SOTA performance among existing methods, with the highest average success rate of 82.6\% and the best average ranking of 1.9 across all tasks.
Specifically, our TUDP attains the best performance on 8 tasks and the suboptimal performance on 6 tasks.
Moreover, compared with the diffusion-based 3D Diffuser Actor, our TUDP significantly improves success rates on tasks with multiple successful actions.
Performance degradation on some tasks is probably caused by multi-task training.

\textbf{Success Rate in a Single-view Setting.}
Following the setting of GNFactor~\cite{ze2023gnfactor}, we also conduct experiments on 10 tasks~(a subset of the 18 tasks) with a single view, the front camera.
Since the single-view scenario has received less attention, only a few existing works have conducted performance experiments in this setting.
As shown in Table~\ref{Evaluation on single view camera}, our TUDP achieves the highest success rate of 83.8\% and the best average ranking of 1.4-th.
Specifically, TUDP has performance improvements on 5 tasks, including \textit{meat off grill, turn tap, put in drawer, push buttons, stack blocks}.
These experiments verify the effectiveness of our method in single-view robotic manipulation task scenarios.

\textbf{Success Rate at Different Iterations.}
Based on the time-unified velocity field, our TUDP generates robot actions efficiently and accurately.
In the other word, our method can also successfully complete the manipulation tasks under the constraint of short generation time.
As shown in Figure~\ref{comparison of efficiency and accuracy}, we set denoising iterations to control generation time and compare the performance of our method and the baseline 3D diffuser actor~\cite{ke2024d}.
When deonising actions with fewer iterations, the average success rate of our TUDP decreases less than that of the 3D Diffuser Actor.
Our method is able to generate robot actions with higher quality in very few iterations, which represents near real-time responsiveness.
Moreover, the velocity limitation $v=0.5$ reduces the efficiency of iterative denoising, resulting in suboptimal performance at low iterations.

\begin{figure}[t]
\centering
\fontsize{8.5}{12}\selectfont
\includegraphics[width=8.5cm]{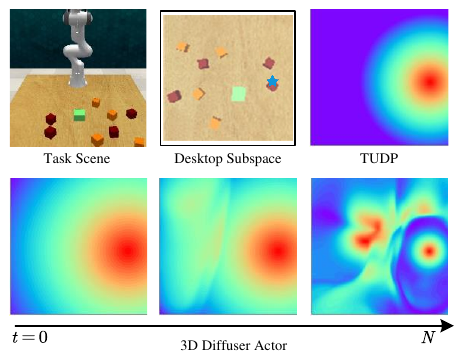} 
\caption{\textbf{Visualization of the magnitude in different velocity fields.} 
In the same scene, we compare the time-unified velocity field of TUDP with the time-varying velocity field of the existing method.}
\label{visualize}
\end{figure}

\textbf{Visualization of Velocity Fields.}
In order to analyze the effectiveness of our method in detail, we visualize the velocity fields of a complex scene in Figure~\ref{visualize}, which aims to \textit{stack maroon blocks}.
We capture the action space along the desktop and mark the label action with a blue star.
We use the magnitude of noises to intuitively display the velocity fields, where small noises are in red while large noises are in blue.
TUDP is compared with the 3D Diffuser Actor, which constructs a time-varying velocity field.
In addition, the time-varying velocity field is not precise enough at $t = 0$ and has a small accurate region at $t = N$.
Meanwhile, the time-unified velocity field of TUDP with velocity limitation achieves accurate iterative denoising.
Thanks to the above characteristics, our method can denoise any noisy action into a successful action.

\begin{table}[t]
    \centering
    \fontsize{8.5}{12}\selectfont
    \begin{tabular}{ccc|cc} 
        \toprule
             \makecell{Action-wise\\[-6pt] Training} & \makecell{Velocity\\[-6pt] Limitation} & \makecell{Early\\[-6pt] Termination}&\makecell{Avg.\\[-6pt] Success.}&  \makecell{Inference\\[-6pt] Time}\\ 
        \midrule
         \textcolor{black}{\ding{51}} &  \textcolor{black}{\ding{51}} &  \textcolor{black}{\ding{51}} & \textbf{82.6} & 0.43 \\ 
         \textcolor{black}{\ding{55}} &  \textcolor{black}{\ding{51}}&  \textcolor{black}{\ding{51}}& 79.1 & 0.43 \\ 
         \textcolor{black}{\ding{51}}&  \textcolor{black}{\ding{55}} & \textcolor{black}{\ding{51}}& 80.9 & \textbf{0.38} \\ 
         \textcolor{black}{\ding{51}}& \textcolor{black}{\ding{51}}& \textcolor{black}{\ding{55}}& 81.8& 0.48  \\ 
         \textcolor{black}{\ding{51}}& \textcolor{black}{\ding{55}}& \textcolor{black}{\ding{55}}& 81.0& 0.45\\
        \bottomrule
    \end{tabular}
    \caption{\textbf{Evaluations on components of our method.}
    We measure effectiveness by average success rate and action generation time during inference.}
    \label{ablation1}
\end{table}

\renewcommand\thesubtable{\alph{subtable}}

\begin{table}[t]
    \centering
    \fontsize{8.5}{12}\selectfont
    \begin{tabular}{c|cccc}
        \hline
        Neighborhood Radius & 0.4 & 0.2 & 0.1 & 0.05 \\
        \hline
        Avg. Success. & 76.5 & 81.4 & \textbf{82.0} & 81.8 \\
        \hline
    \end{tabular}
    \caption{\textbf{Neighborhood radius of successful actions.}
    We choose an appropriate neighborhood radius to adapt to the distribution of successful actions.}
    \label{ablation2}
\end{table}

\begin{table}[t]
    \centering
    \fontsize{8.5}{12}\selectfont
    \begin{tabular}{c|ccccc}
        \hline
        Gaussian Variance &  1 & 0.5 & 0.25 & 0.15 & 0.05 \\
        \hline
        Avg. Success. &  78.6 & \textbf{81.9}  & 81.8 & 80.7 & 78.3\\
        \hline
    \end{tabular}
    \caption{\textbf{Gaussian variance of noisy actions.}
    We choose an appropriate Gaussian variance to take into account both global and local denoising.}
    \label{ablation3}
\end{table}

\subsection{Ablation Experiments}

\textbf{Ablation on the Components of TUDP.}
We have verified the effectiveness of important components in the method by comparing the average accuracy of the tasks in an ablation experiment.
As shown in Table~\ref{ablation1}, we use crosses to indicate components that are currently abandoned.
The action-wise training method is fundamental for our TUDP.
Due to the importance of action discrimination ability in action denoising, canceling action-wise training methods leads to a significant decrease in the success rate in row 2.
Comparing row 1 and row 3, the velocity limitation improves the accuracy of noise prediction to a certain extent, but also spends unnecessary time on simple tasks.
The early termination in the iterative denoising process has no significant impact on the average success rate in row 4 and shortens the iterative denoising process, especially when velocity limitation is abandoned as shown in row 5.

\textbf{Ablation on the Hyperparameters of TUDP.}
We verified the rationality of the hyperparameter selection in the model through experiments.
We selected the appropriate neighborhood radius of successful actions and Gaussian variance of noisy actions through ablation experiments and analyzed their effects.
As shown in Table~\ref{ablation2}, a huge neighborhood radius can confuse different successful actions, while a neighborhood radius for elementary schools can increase the learning difficulty of action discrimination networks.
And the appropriate neighborhood radius $l=0.1$ achieves the highest average success rate.
As shown in Table~\ref{ablation3}, Gaussian variance $\sigma=1$ has the lowest average successful rate, because smaller Gaussian variances force the policy network to focus on successful actions.
To take into account the random noisy actions in action space at the same time, appropriate variance $\sigma=0.5$ achieves a better performance on robotic manipulation.

\begin{figure}[t]
    \centering
    \fontsize{8.5}{12}\selectfont
    \includegraphics[width=12cm]{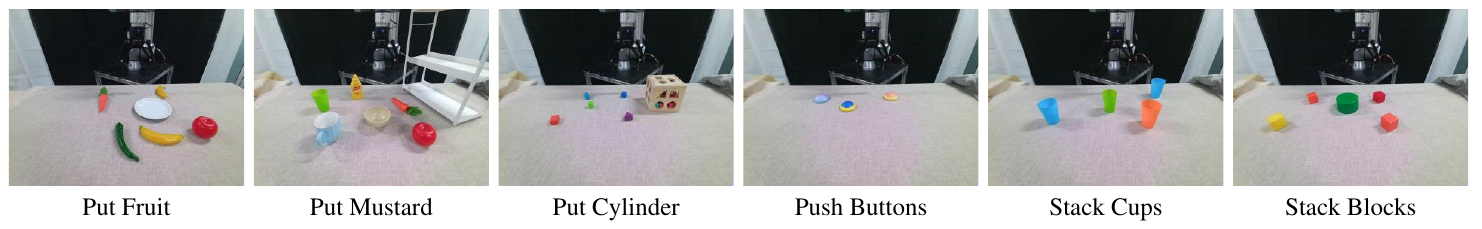} 
    \caption{\textbf{Overview of 6 real-world tasks.} We show the image observations in these real-world scenarios.}
    \label{real-world demo}
\end{figure}

\begin{table}[h]
    \centering
    \fontsize{8.5}{12}\selectfont
    \begin{tabular}{c|cc|c}
        \hline
        Tasks & Variations & Training Demos & Test Success ($\uparrow$) \\
        \hline
        Push Buttons & color and object(2)& 10& 10/10 \\
        Put Fruit in bowl & object(4)&10 & 10/10 \\
        Put Mustard on shelf & object(2)& 10&  9/10 \\
        Put Cylinder into box &color(2) &8 & 8/10\\
        Stack Blocks &color(2) & 12& 8/10 \\
        Stack Cups &N/A(1)& 8& 6/10\\
        \hline
        All Tasks &13 & 58& 51/60\\ 
        \hline
    \end{tabular}
    \caption{\textbf{Real-world tasks configuration and performance.} We conduct experiments on 6 real-world tasks, evaluating 10 episodes for each task and reporting the success rate.}
    \label{real-world performance}
\end{table}

\subsection{Real-Robot Results}

We perform a series of experiments with a 6-DoF UR5 robotic arm, which is fitted with a Robotic 2F-140 two-finger gripper and a  single Azure Kinect (RGB-D) camera in the front view. 
Our real-robot setup encompasses six language-conditioned tasks, featuring several variations, including various objects and colors. 
In total, we gathered 58 demonstrations. 
We visualize different regions influenced by the dynamic radius schedule in the real world, following trajectories defined by human-specified waypoints. 
For inference, we apply the BiRRT planner integrated with MoveIt! ROS package~\cite{coleman2014reducing} to act towards the predicted robot poses. 
Visual examples of our tasks can be found in Figure~\ref{real-world demo}, with quantitative results presented in Table~\ref{real-world performance}. 
We have also achieved satisfactory results on difficult tasks with high precision requirements, such as \textit{put cylinder into box} and \textit{stack blocks}.
These results show that TUDP effectively learns precise, real-world manipulation tasks from a limited set of demonstrations and achieves reliable task execution.

\section{Limitation and Conclusion}
\label{Conclusion}

\textbf{Limitation.}
The outstanding performance of the Time Unifie Diffusion Policy~(TUDP) depends on the robotic manipulation tasks satisfying certain prior assumptions.
In particular, the neighborhood radius of successful actions hinders the application of our method in dense interaction scenarios.
Meanwhile, our TUDP uses several hyperparameters to adjust the action denoising, which leads to a higher complexity of model tuning.

\textbf{Conclusion.}
In this study, we propose TUDP for robotic manipulation that maintains both high accuracy and efficiency. 
TUDP establishes a time-unified velocity field with successful action discrimination, significantly reducing action denoising time without compromising performance. 
Our action-wise training method facilitates TUDP in learning the time-unified velocity field with implicit discrimination information of successful action.
Empirical results show that TUDP outperforms existing methods, achieving a state-of-the-art average success rate on the RLBench benchmark.
More importantly, under the restriction of low denoising iterations, our TUDP shows a more significant performance advantage.

\section*{Acknowledgment}
This work was supported in part by the National Science and Technology Major Project under Grant 2023ZD0121300, in part by the National Natural Science Foundation of China under Grant 62088102, Grant U24A20325, and Grant 12326608, and in part by the Fundamental Research Funds for the Central Universities under Grant XTR042021005.

\bibliographystyle{elsarticle-num}
\bibliography{ref.bib}


\end{document}